\documentclass[letterpaper, 10 pt, conference]{ieeeconf}  

\IEEEoverridecommandlockouts                              

\usepackage{booktabs}
\usepackage[pdfborder={0 0 0},citecolor={blue},colorlinks={True},linkcolor={blue}]{hyperref}
\usepackage{subcaption}
\usepackage{graphicx}
\newcommand{\mycomment}[1]{}
\usepackage{lipsum}

\usepackage{times}
\usepackage{balance}
\usepackage{cite}
\usepackage{amsmath,amssymb,amsfonts}
\usepackage[ruled,vlined,linesnumbered]{algorithm2e}
\SetAlFnt{\small}
\usepackage{algpseudocode}
\usepackage{textcomp}
\usepackage{mathtools}
\usepackage{enumerate}
\usepackage{bm}
\usepackage{makecell,multirow,diagbox}
\usepackage{xcolor}
\usepackage{threeparttable}
\usepackage{booktabs}
\usepackage[symbol]{footmisc}
\usepackage{lipsum}  

\title{\LARGE
NeuRSS: Enhancing AUV Localization and Bathymetric Mapping with Neural Rendering for Sidescan SLAM}

\author{Yiping Xie$^{1}$, Jun Zhang$^{2}$, Nils Bore$^{3}$ and John Folkesson$^{1}$
\thanks{*This work was supported in part by the Wallenberg AI, Autonomous Systems and Software Program (WASP) funded by the Knut and Alice Wallenberg Foundation and in part by Stiftelsen för Strategisk Forskning (SSF) through the Swedish Maritime Robotics Centre (SMaRC) under Grant IRC15-0046.
(Corresponding author: Yiping Xie.)}
\thanks{$^{1}$Yiping Xie and John Folkesson are with Robotics, Perception and Learning Division, KTH Royal Institute of Technology, Stockholm, Sweden (e-mail: \{yipingx, johnf\}@kth.se).}
\thanks{$^{2}$Jun Zhang is with the Institute of Computer Graphics and Vision, Graz University of Technology, Graz, Austria (email: jun.zhang@tugraz.at).}
\thanks{$^{3}$Nils Bore is with Ocean Infinity (email: nils.bore@oceaninfinity.com).}}
\begin{document}

\maketitle

\thispagestyle{empty}
\pagestyle{empty}

\begin{abstract}
Implicit neural representations and neural rendering have gained increasing attention for bathymetry estimation from sidescan sonar (SSS). These methods incorporate  multiple observations of the same place from SSS data to constrain the elevation estimate, converging to a globally-consistent bathymetric model. However, the quality and precision of the bathymetric estimate are limited by the positioning accuracy of the autonomous underwater vehicle (AUV) equipped with the sonar. The global positioning estimate of the AUV relying on dead reckoning (DR) has an unbounded error due to the absence of a geo-reference system like GPS underwater. To address this challenge, we propose in this letter a modern and scalable framework, \textit{NeuRSS}, for SSS SLAM based on DR and loop closures (LCs) over large timescales, with an elevation prior provided by the bathymetric estimate using neural rendering from SSS. This framework is an iterative procedure that improves localization and bathymetric mapping. Initially, the bathymetry estimated from SSS using the DR estimate, though crude, can provide an important elevation prior in the \textit{nonlinear least-squares} (NLS) optimization that estimates the relative pose between two loop-closure vertices in a pose graph. Subsequently, the global pose estimate from the SLAM component improves the positioning estimate of the vehicle, thus improving the bathymetry estimation.
We validate our localization and mapping approach on two
large surveys collected with a surface vessel and an AUV, respectively. We evaluate their localization results against the ground truth and compare the bathymetry estimation against data collected with multibeam echo sounders (MBES).
\end{abstract}

\section{INTRODUCTION}
Small autonomous underwater vehicles (sAUVs) equipped with sidescan sonar (SSS) are often used for hydrogeological surveys and seabed mapping. 
Nonetheless, in the absence of GPS and an \textit{a-priori} map of the surveyed area, the dead reckoning (DR) estimate of their global position can drift severely over time. Existing underwater positioning systems analogous to GPS, such as long baseline (LBL) and ultra-short baseline (USBL), require external infrastructure for the deployment of beacons/transponders. As a small-form and low-cost sensor, SSS provides a promising and cost-effective solution for sAUV navigation and mapping, due to its ability of generating high-resolution images with wide swath. 

Traditionally, bathymetric maps are usually constructed with multibeam echo sounders (MBES), which can be cost-prohibitive and too large for low-cost sAUVs. For SSS data, the range and the azimuth angle of the returns are known, but the information of the elevation angle is lost due to projection, which is essential for bathymetry reconstruction. However, since the changes of the returned intensities indicate changes of the incidence angle, it is possible to extract information of the slope of the seafloor from SSS data. Although reconstruction of the seafloor from single line of SSS imagery is mathematically ill-posed, this problem can be constrained adequately in areas that have been observed from multiple viewpoints. One can notice the similarities between bathymetry reconstruction from SSS and 3D reconstruction from camera images, in the sense that for optical camera images, there is also one dimensional information lost during the projection, that is, the range instead of the elevation angle. As a result, many approaches from computer vision and computer graphics for 3D reconstruction using camera images can be adapted for the same task using sonars. Examples of these are, in the early days, shape-from-shading (SfS) techniques~\cite{li1991improvement,lambertian1991, johnson1996seafloor, coiras2007tip}, and more recently data-driven methods using convolutional neural networks (CNNs)~\cite{xie_cnn,xie_neural_normal} and inverse rendering based on implicit neural representations~\cite{bore2022neural,xie2022sidescan, qadri2023neural, xie2024FLS}.

\begin{figure}[!h]
\centering
\includegraphics[width=\linewidth]{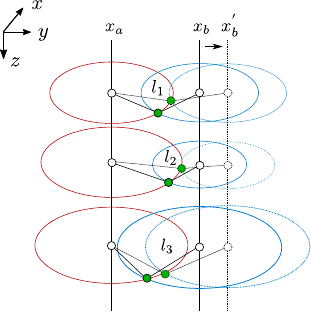}  
  \caption{Illustration of landmark elevation degeneracy with pure $y\textrm{-}$translation motion, in the sensor forward-lateral-down frame. Here we show two submaps from two parallel survey lines with 3 landmarks (green dots). In the NLS optimization, we usually 
  fix pose $\bm{x}_a$, namely fixing the solid red circles (indicating the range measurements). Without any priors on landmarks, $\bm{x}_b$ and all the landmarks can move together in $y\textrm{-}z$ plane, in this case, $\bm{x}_b$ positive translation along the $y\textrm{-}$axis to $\bm{x}_b^{'}$, landmarks moving up along the $z\textrm{-}$axis (from solid strokes to dashed strokes), where all SSS range and bearing measurements are still fulfilled. }
  \label{fig:elevation_degeneracy}
\end{figure}

As for underwater navigation, simultaneous localization and mapping (SLAM) techniques can reduce the drift in the AUV's DR estimate using onboard sensor measurements. Most of the state-of-the-art graph-based underwater SLAM solutions for unstructured environment target AUVs equipped with MBES~\cite{torroba2020pointnetkl}. But the limited coverage of MBES makes loop closure (LC) detection sparse, especially due to the scarcity of distinguishable features on the seabed. SSS, on the other hand, has wider coverage which potentially allows more LC detections between overlapping submaps constructed from adjacent survey lines, and can potentially  resolve smaller features.
However, due to the elevation ambiguity inherent to SSS sensor, graph-based SSS SLAM faces elevation degeneracy (illustrated in Fig.~\ref{fig:elevation_degeneracy}), making the overall NLS optimization rank-deficient. Such degeneracy, similar to forward looking sonar (FLS)~\cite{huang2015towards, huang2016incremental, westman2018icra, westman2019joe,FLS_SLAM_terrain}, results in large errors in landmark
estimation and unrobustness in the relative pose estimation between two LC vertices of a pose graph when a LC is detected. The degeneracy case is especially common in standard lawn-mower pattern\footnote{A lawn-mower pattern that has the vehicle perform the survey as a series of long parallel lines.} surveys, since most matches can only be made between parallel lines where the SSS sees the feature from the same side but different distances.

This degeneracy is a well-known issue in triangulation, as a result, most SSS  SLAM approaches address such elevation degeneracy by assuming the seafloor is locally flat~\cite{fallon2011icra,bernicola2014oceans,issartel2017oceans}, and as for FLS SLAM approaches, they use the planar assumption~\cite{xu2022robust}. This assumption only works when the structure of seafloor is relatively benign, for example, slowly sloping. The assumption will break when dealing with complex seabed such as rocky areas, mountains and ridges. However, in these scenarios, an estimate of the seafloor would be of great help to constraining the elevation angle of the returns, serving as an elevation prior in the NLS optimization, yielding more robust relative pose estimation. The easiest way to estimate the rough bathymetry is to linearly interpolate altimeter readings so that a crude bathymetric model can be obtained, but the errors are still quite large if the seabed is complex. On the other hand,  incorporating information from SSS data would significantly improve the bathymetric model.

Bathymetry estimation from SSS requires accurate pose estimation, while SSS SLAM suffers from the elevation degeneracy without precise bathymetric estimate. Nevertheless, one can address such a chicken-and-egg problem by iteratively improving the estimates of the bathymetry and the pose estimation. Following this principle, we present a framework\footnote{The source code link will be provided here upon acceptance.}, \textit{NeuRSS}, that leverages the advances of neural rendering to estimate bathymetry from SSS~\cite{bore2022neural, xie2022sidescan}, which provides an elevation prior addressing the degeneracy in SSS SLAM problems and showcase that one can significantly improve the AUV's DR estimate and subsequently produce high-quality bathymetry using SSS data from a standard survey.

The contribution of this work is to extend our previous work~\cite{bore2022neural, xie2022sidescan,zhang2023ietrsn} and combine them into a new framework. For the sonar scattering model in~\cite{xie2022sidescan}, we extend it to be able to model shadows. For the back-end of the SSS SLAM framework in~\cite{zhang2023ietrsn}, we extend it to a submap-based optimization with a prior map to better address the elevation degeneracy. We evaluate the proposed NeuRSS framework numerically using two field datasets containing both SSS and MBES collected
simultaneously.

\section{RELATED WORK}

\subsection{SSS SLAM}
Early works used stochastic maps to estimate the AUV's position from SSS imagery with an extended Kalman filter (EKF)~\cite{ruiz2003oceans,ruiz2004concurrent,reed2006tip}, however, EKF-based SLAM approaches suffer from scalability as size of  the state vector grows, especially in a large underwater environment, e.g., open sea.
Later, Fallon~\textit{et al.}~\cite{fallon2011icra} proposed a graph-based SLAM framework that utilizes incremental smoothing and mapping (iSAM) to fuse acoustic ranging and SSS measurements for on-board applications in real time.
Similarly, Bernicola~\textit{et al.}~\cite{bernicola2014oceans} demonstrated using iSAM to correct trajectories with SSS images.
Issartel~\textit{et al.}~\cite{issartel2017oceans}, aiming to address the false SSS data associations issue, proposed to
 use switchable observation constraints in the pose graph. However,  they~\cite{fallon2011icra,bernicola2014oceans,issartel2017oceans} all use the flat seafloor assumption for the sonar measurement model, which prevents the NLS optimization from landmark elevation degeneracy (pure y-translation). But the error introduced by this assumption increases as the terrain gets complex, harming the performance of localization.
In~\cite{zhang2023ietrsn,zhang2023dense} the 3D landmark positions together with relative pose transformation between the reference pose and current pose are estimated in the NLS optimization with an elevation prior on the landmark, provided by interpolating between the altitudes of the two poses. However, the underlying assumption is that the terrain is relatively benign. 

\subsection{FLS SLAM}
Another line of work that is highly relevant is FLS SLAM~\cite{huang2015towards, huang2016incremental, westman2018icra, westman2019joe,FLS_SLAM_terrain}, where the degeneracy cases are in general more complex than SSS SLAM. Different robot motions
on this degeneracy are discussed in~\cite{huang2015towards}, where the majority of the cases causing such does not exist in SSS SLAM with a standard lawn-mower pattern, except the pure y-translation. 
Westman~\textit{et al.}~\cite{westman2018icra,westman2019joe} proposed to examine the eigenvalues of the Jacobian
matrix of landmarks to address this degeneracy cases whereas Wang~\textit{et al.}~\cite{FLS_SLAM_terrain} modeled the terrain as Gaussian Processes (GPs) and incorporated the terrain factors into the factor graph.


\subsection{SSS Bathymetry Reconstruction}
Early works to estimate bathymetry from SSS rely on traditional SfS techniques and Lambertian models. Notably, Coiras~\textit{et al.}~\cite{coiras2007multiresolution} used a Lambertian model for the sonar ensonification process and obtained an approximation of the surface gradients from sonar intensities by inverting the
image formation. They showed that convergence can be improved by gradually increasing the resolution of the predicted bathymetry. Recently, data-driven approaches~\cite{xie_cnn,xie_neural_normal} using CNNs have been proposed to learn the missing elevation directly from SSS images in a supervised-learning fashion. However, the ``ground truth'' bathymetry is needed for creating a training set, which is not always practical underwater. Neural rendering~\cite{bore2022neural,xie2022sidescan} methods that leverage the continuity and differentiablity of implicit neural representations have been recently proposed to fit many sidescan lines into a self-consistent bathymetry with a global optimization. Specifically, a multi-layer perceptron (MLP) with sine activation functions, known as SIREN~\cite{siren2020} was used to represent the bathymetry where the gradients of the bathymetry were constrained by SSS intensities through a Lambertian model. Extended from~\cite{bore2022neural}, a nadir model was proposed in~\cite{xie2022sidescan} to model the nadir region in SSS waterfall images so that the optimization can converge without any external bathymetric data, e.g., altimeter readings. However, acoustic shadows cannot be explained by the Lambertian model in~\cite{xie2022sidescan}. Furthermore, all the aforementioned works assume to have access to high-accuracy navigation estimates.

\section{Neural Rendering for Bathymetry Estimation}\label{sec:NR}
A prerequisite of our neural rendering pipeline is the assuming that the vehicle’s trajectory is already corrected. In this section, we present the neural rendering pipeline with an extended Lambertian model, based on implicit neural representations.

\subsection{Implicit Neural Representation}
In this approach, the bathymetry is represented using an implicit neural representation, specifically a function $\Phi_\theta:\mathbb{R}^2 \rightarrow \mathbb{R}$, which maps 2D spatial coordinates, i.e., Euclidean easting and northing $x,y$, to the corresponding height of the seafloor $\tilde{h}$. This function is parameterized by a fully connected neural network with parameters $\theta$, specifically, a variant of a MLP that employs sinusoidal activation functions, known as SIREN~\cite{siren2020}. 

Given a SSS survey in a dataset of the form $D=\{I_i,\bm{x}_i,h_i\}_{i=1}^{N}$, containing $N$ pings of SSS intensities $I_i \in \mathcal{I}$, estimates of the 6D AUV poses $\bm{x}_i$ and altimeter readings $h_i$. Combining the positioning estimates of the AUV and $h_i$, we have sparse bathymetric measurements on the seafloor surface $\{p_i^{xyz}\}_{i=1}^{N}$ along the AUV trajectory in the world coordinates ENU (Easting, Northing, Up), which can be directly used to constrain the unknown mapping $\Phi_\theta$:
\begin{equation}
    \mathcal{L}_H=\frac{1}{\lvert \{p^{xyz}_i\}\rvert}\sum_i \lVert \Delta^{\Phi_\theta}(p^{xyz}_i) \rVert,
\end{equation}
where 
\begin{equation}
\Delta^{\Phi_\theta}(p)=\Phi_\theta(p_x,p_y)-p_z
\end{equation}
is the signed vertical distance to the seafloor, $\lvert \cdot \rvert$ denotes the size of the set $\{p^{xyz}_i\}$ and $\lVert \cdot \rVert$ denotes the $L_2$ norm. 

Besides the constraint $\mathcal{L}_H$ from the altimeter readings, the measured returned intensity of every pixel in the SSS images $I_{i,n}$ at given ping $i$ and given bin $n$ can be modelled using neural rendering given a sidescan scattering model. The difference between the rendered SSS intensity $\tilde{I}_{i,n}$ and the measured intensity $I_{i,n}$ can form the intensity loss to further constrain the surface normal of the bathymetry $\Phi_\theta$:
\begin{equation}
    \mathcal{L}_{I}=\frac{1}{\lvert \{I_{i,n}\}\rvert}\sum_{I_{i,n}} \lVert \tilde{I}_{i,n}-I_{i,n} \rVert.
\end{equation}
The following subsection introduces the sidescan scattering model and the process of neural rendering.

\subsection{Sidescan Scattering Model}
\begin{figure}[t]
\centering
\includegraphics[width=\linewidth]{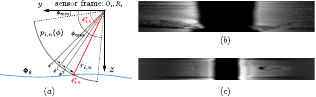}  
  \caption{(a) Illustration of the gradient descent approach to find the intersection between the elevation arc and the seafloor, parameterized by SIREN. (b) An example of a SSS image in Dataset 1. (c) An example of a SSS image in Dataset 2, showing the sinkhole on the seabed.}
  \label{fig:GD}
\end{figure}
SSS emits a fan-shape beam to the side of the AUV with a narrow beam along the travel direction and a wide beam in the azimuth direction, and then records the returned echos at fixed intervals of time. The recorded backscatter intensities are arranged in a vector, often referred as a \textit{ping}, which can be stacked ``row-by-row'' as the vehicle moves along to form a ``waterfall'' image [see Fig.~\ref{fig:GD} (b) and (c)]. Each item in the vector, often referred as a \textit{bin}, stores the amplitude and two-way travel time of the returns. The travel time is used to calculate the distance of returns from the sonar array, combined with the sound velocity profile (SVP). 

Similar to~\cite{nils_nsfs,xie2022sidescan}, we use the Lambertian model for the scattering process. For $I_{i,n}$, denoting the measured returned intensity from the ensonified point on the seafloor $p_{i, n}$, its returned intensity can be approximated by:
\begin{equation}\label{eq:Lambertian}
    \tilde{I}_{i,n}=K \Phi(p_{i, n})  R(p_{i, n})  \left\Vert \cos(\alpha) \right\Vert^2,
\end{equation}
where $K$ is the normalizing constant, $\Phi$ is the beam pattern of the sonar, $R$ is the reflectivity of the seafloor and $\alpha$ is the incidence angle. The incidence angle can be calculated given sonar's pose and the bathymetry model $\Phi_\theta$ as following. 

Assuming isovelocity SVP, the ensonified volume $p_{i, n}$ is at a fixed distance (slant range) $r_{i,n}^s$ away from the sonar, parameterized by the elevation angle $\phi_{i,n}^g$ along an arc referred to as an \textit{isotemporal curve}. Given the estimated bathymetry $\Phi_\theta$, we can determine the elevation angle $\phi_{i,n}^g$ by using gradient descent (GD) algorithm to find where the arc intersects with the current estimated seafloor (see Fig. \ref{fig:GD}), assuming the arc only has one intersection with the seafloor within sonar's vertical sensor opening $[\phi_{\textrm{min}},\phi_{\textrm{max}}]$:
\begin{equation}
    \phi^{k+1}=\phi^k - \frac{\lambda}{r_{i,n}}\frac{d}{d\phi}(\Delta^{\Phi_\theta}(p_{i,n}(\phi^k)))^2,
\end{equation}
where $\Delta^{\Phi_\theta}(p)$ is the signed vertical distance to the seafloor, and $\lambda$ is the step size for gradient descent.
Once we find the optimal $\phi_{i,n}^g$,  we can define the surface normal at $p_{i, n}$ with respect to $\Phi_\theta$, given the gradient components $\nabla_x$,$\nabla_y$:
\begin{equation}
 N^{\Phi_\theta}(p) = \left[ -\nabla_x \Phi_\theta(p_x, p_y), -\nabla_y \Phi_\theta(p_x, p_y), 1 \right]^T.
\end{equation}
The ray from sonar to the isotemporal curve can be defined as~\cite{nils_nsfs}:
\begin{equation}
 r(\phi_{i,n}^g) = r_{i,n}^s R_i \left[0, -\cos(\phi_{i,n}^g), \sin(\phi_{i,n}^g) \right]^T.
\end{equation}
Given $N^{\Phi_\theta}(p)$ and $r(\phi_{i,n}^g)$, we can compute the Lambertian 
scattering contribution, $M_{i,n}^{\Phi_\theta}$, using $\cos^2$ of the incidence angle:
\begin{equation}
M_{i,n}^{\Phi_\theta} = \left\Vert \cos(\alpha) \right\Vert^2=\bigg (r(\phi_{i,n}^g)^T \hat{N}^{\Phi_\theta}(p_{i,n})\bigg)^2.
\end{equation}
Besides the Lambertian contribution in Eq.~\ref{eq:Lambertian}, we also model and estimate the gain, beam pattern and reflectivity jointly with the bathymetry, similarly as~\cite{bore2022neural,xie2022sidescan}. The beam pattern $\Phi(\phi)$ is modelled as a radial basis function (RBF) with kernels evenly spread across $\phi$ at fixed positions. Reflectivity $R(p)$ of the whole surveyed area is also modelled as a 2D RBF with kernels spread spatially. The gain parameter $A_i$ is estimated for each sidescan line.
\subsection{Nadir and Shadows}
Neither nadir area in the SSS data nor the shadows can be explained by the Lambertian scattering model, thus we propose to extend the traditional Lambertian model to handle both components in the SSS data, inspired by recent advances of volumetric rendering using neural implicit surfaces~\cite{nerf2020, wang2021neus}.

Nadir area is when the sound pulse travels through the water column before hitting the seafloor, where the corresponding arc has no intersections with the seafloor. Since we perform a fixed-number of steps gradient descent to calculate where the arc intersects with the seafloor, for the nadir area, the signed vertical distance $\Delta^{\Phi_\theta}(p(\phi^*))$ for the optimal grazing angle $\phi^*$ would be far from zero. We propose to weight the intensity for nadir area with its volume density, computed as~\cite{xie2022sidescan}: $\sigma(p(\phi^*))=\exp{(-\frac{\Delta^{\Phi_\theta}(p(\phi^*))^2)}{\sigma_s}}$, so that the volume density at nadir area is close to zero but one other wise. $\sigma_s$ is a spread parameter that can be manually tuned to control the smoothness of the volume density function.

It is also well-known that the shadows in the sidescan data where cannot be explained by the Lambertian model. Similarly, for the shadows in the data, the gradient descent procedure would find a intersection between the arc and the seafloor, but the found intersection is occluded. Inspired by~\cite{qadri2023neural,xie2024FLS}, we propose to sample a few points along the ray backwards, starting from the intersection and compute the accumulated transmittance at the intersection $T(p(\phi^*))=\exp{(-\int_0^{r^s}\rho(u)du)}$, which is used to indicate if $p(\phi^*)$ is occluded. Here, the particle density $\rho$ is computed as in~\cite{qadri2023neural,xie2024FLS}. We use $T(p(\phi^*))$ to weight the predicted SSS intensity, since if $T(p(\phi^*))$ is close to zero, it indicates that the intersection point is occluded, which corresponds to shadows in the sidescan images.
The extended Lambertian model for the complete intensity rendering is given by
\begin{equation}
    \tilde{I}_{i,n}=A_i \Phi(\phi^*_{i,n}) R(\phi^*_{i,n}) M_{i,n}^{\Phi_\theta}(\phi^*_{i,n}) \sigma^{\Phi_\theta}(\phi^*_{i,n}) T^{\Phi_\theta}(\phi^*_{i,n}).
\end{equation}
\subsection{Neural Rendering Algorithm}

Algorithm~\ref{alg:NR} outlines the SIREN training process using neural rendering. The inputs are the dataset $D$ collected during the survey and the learned parameters are $\{\Phi_\theta, \Phi, R, A_i\}$ containing the bathymetry estimation $\Phi_\theta$ parameterized by SIREN, the beam pattern $\Phi$ parameterized by 1D Gaussian kernels, the reflectivity of the seafloor $R$ parameterized by 2D Gaussian kernels and gains per sidescan lines $A_i$ as scalar variables. The initialization of $\Phi_\theta$ can be random as in~\cite{nils_nsfs,xie2022sidescan} or one could apply bilinear interpolation given $\{\bm{x}_i,h_i\}$ to obtain an initial estimate of $\Phi_\theta$, which in theory would fasten the convergence of the SIREN training afterwards. Lines $2-15$ outline the SIREN training for a fixed number of steps ($K$) with a decaying learning rate. Specifically, lines $3-6$ calculate the height loss within a batch of random samples with batch size $M_H$ given the current bathymetry estimate $\Phi_{\theta}^k$. Lines $8-13$ calculate the intensity loss within a batch of data (batch size $M_I$), where line $9$ applies a fixed number of steps GD to find the optimal elevation angle $\phi$ for each sample given the current bathymetry estimate $\Phi_{\theta}^k$. Line $10$ computes the beam pattern and reflectivity at $\phi$ given the current estimate of $\Phi^k, R^k$ and line $11$ calculates the corresponding Lambertian component $M$, the volume density $\sigma$ and the accumulated transmittance $T$ of the seafloor intersection. Lines $12-13$ construct the intensity loss within the batch and line $14$ updates $\{\Phi_\theta, \Phi, R, A_i\}$ at the same time using gradient-based optimization.

\begin{algorithm}
 \KwData{$D=\{I_{i,n},\hat{\bm{x}}_i,h_i\}$}
 \KwResult{$\Phi_\theta, \Phi, R, A_i$ }
 initialization\\
 \For{$k=0$ to $K-1$}{
 random sample $M_H$ samples from $\{\hat{\bm{x}}_i,h_i\}$\\
 \For{$m_H=0$ to $M_H$}{$p_{m_H}^{xyz} \gets \hat{\bm{x}}_{m_H}, h_{m_H}$\\
 $\mathcal{L}_H \gets p_{m_H}^{xyz},\Phi_{\theta}^k $
 } 
 random sample $M_I$ samples from $\{\hat{\bm{x}}_i,I_{i,n}\}$\\
 \For{$m_I=0$ to $M_I$}{
 $\phi_{m_I} \gets \textrm{GD}(\hat{\bm{x}}_{m_I}, \Phi_{\theta}^k, r_{m_I}$)\\ 
 $\Phi_{m_I}^k,R_{m_I}^k\gets \hat{\bm{x}}_{m_I},\phi_{m_I},r_{m_I} $ \\ 
 $M_{m_I},\sigma_{m_I},T_{m_I} \gets \hat{\bm{x}}_{m_I},\phi_{m_I},r_{m_I},\Phi_{\theta}^k $ \\ 
 $\hat{I}_{m_I} \gets  A_{m_I}^k, \Phi_{m_I}^k, R_{m_I}^k, M_{m_I}, \sigma_{m_I}, T_{m_I}$\\
 $\mathcal{L}_I \gets I_{m_I}, \hat{I}_{m_I}$
 } 
 $\Phi_{\theta}^{k+1},\Phi^{k+1},R^{k+1}, A^{k+1} \gets \textrm{OptimizerStep}(\mathcal{L}_H,\mathcal{L}_I, \Phi_{\theta}^{k},\Phi^{k},R^{k}, A^{k})$\\
 update the learning rate
 }
 \Return $\Phi_{\theta}^{K},\Phi^{K},R^{K}, A^{K}$
 \caption{$\Phi_\theta$ Training.}\label{alg:NR}
\end{algorithm}
\section{SSS SLAM with Elevation Prior}\label{sec:slam}
The full 6D AUV state is defined as $[x,y,z,\phi,\theta,\psi ]$ in three Cartesian and three rotation dimensions, where absolute measurements of the depth $z$, roll $\phi$ and pitch $\theta$, are measured using a pressure sensor and inertial sensors, respectively. The heading $\psi$ is measured using a compass and integrated with DVL measurements to propagate estimates of $x$ and $y$, which will drift over time. The motion of the AUV ${\bm{x}}_i$ can be modelled as a Gaussian distribution ${\bm{x}}_i = \mathcal{N}(f({\bm{x}}_{i-1}, {\bm{u}}_i), \Sigma_i)$, where ${\bm{u}}_i$ is the vehicle's control input, $f(\cdot)$ is its motion model and $\Sigma_i$ is the covariance of the additive Gaussian noise that parameterizes the uncertainty of the DR.

\subsection{Sidescan Sonar Measurement Model}
 A landmark on the seafloor ${\bm{l}}$ gives a range measurement, namely the slant range, $r_s$, and a bearing measurement from the constraint that ${\bm{l}}$ lays within the horizontal sensor opening in the $y\textrm{-}z$ plane. These two measurements paired together can be written as 
\begin{eqnarray}
{\bm{z}}_m = \begin{pmatrix} r_m \\ 0\\ \end{pmatrix} = \hat{\bm{z}}_m + {\bm{\eta}}  = \begin{pmatrix} \sqrt{{\bm{\pi}}({\bm{l}}_m)\cdot{\bm{\pi}}({\bm{l}}_m)}  \\ (1,0,0)\cdot{\bm{\pi}}({\bm{l}}_m)\\ \end{pmatrix} + {\bm{\eta}},
\label{eq:KMM}
\end{eqnarray}
where ${\bm{l}}_m\in \mathbb{R}^3$ is the 3D landmark in the world frame, $r_m$ is the slant range of ${\bm{l}}_m$ and $\bm{\eta}$ is the measurement noise. $\bm{\pi}(\cdot)$ is a function that transforms a 3D landmark from world frame to the sensor frame (denoted by $s$):
\begin{eqnarray}
\bm{\pi}(\bm{l})={}^{s}\bar{\bm{l}} = {}^{p}{\bm{T}}_{s}^{-1}\cdot{}{\bm{T}}_{p}^{-1}\cdot\bar{\bm{l}}.
\label{eq:transform}
\end{eqnarray}
Here ${\bm{T}}_{p} \in \mathrm{SE}(3)$ denotes the AUV body pose at current ping (denoted by $p$) that contains the $m^\textrm{th}$ keypoint, and ${}^{p}{\bm{T}}_{s}$ is the transformation from AUV body frame to its sensor frame. $\bar{\bm{l}} \in \mathbb{E}^3$ denotes the homogeneous representation of ${\bm{l}}$.
\subsection{Submap-based Relative Pose Estimation with a Prior Map}
For a submap-based approach, we reasonably assume that the DR error within submaps is small enough to be neglected. Then for Eq.~\ref{eq:transform}, we need one more transformation  from the center pose of the submap to the pose at the ping corresponding to the landmark, ${}^{p}{\bm{T}}_{c}$:
\begin{eqnarray}
\bm{\pi}(\bm{l})={}^{s}\bar{\bm{l}} = {}^{c}{\bm{T}}_{s}^{-1}\cdot{}{}^{p}{\bm{T}}_{c}^{-1}\cdot{}{\bm{T}}_{p}^{-1}\cdot\bar{\bm{l}},
\label{eq:transform-submap}
\end{eqnarray}
where ${}^{c}{\bm{T}}_{s}$ now is the sensor offset, which is usually considered as known. Now we can describe the two-view submap-based sonar optimization as the following. 
\begin{figure}[!h]
\centering
\includegraphics[width=\linewidth]{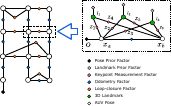}  
  \caption{Factor graph formulation of the proposed framework. Left: Pose graph of the global optimization. Right: factor graph for submap-based relative pose estimation with the elevation prior.}
  \label{fig:factor_graph}
\end{figure}
Given a submap, $S_a$, consisting of $N_a$ sidescan pings which has an overlapping area (containing $M_s$ landmarks) with another submap, $S_b$ with $N_b$ pings, we denote the poses of the center ping for $S_a$ and $S_b$ as $\bm{x}_a$ and $\bm{x}_b$, respectively. 
We formulate this optimization with a-priori map as a factor graph [see Fig.~\ref{fig:factor_graph} (right)], and estimate poses of the center of the two submaps as well as $M_s$ 3D point landmarks, $X=[\bm{x}_a,\bm{x}_b, \bm{l}_1, \bm{l}_2,\cdots, \bm{l}_{M_s}]$, by solving a \textit{nonlinear least-squares} (NLS) problem under Gaussian noise with a \textit{maximum a posteriori} (MAP) estimator:
\begin{align} 
\begin{split}
    X^{*}&=\underset{X}{\textrm{argmin}} \sum_{m=1}^{M_s}\lVert \hat{{\bm{z}}}_m - h(\bm{x}_a,\bm{l}_m)\rVert^2_{\Sigma_{a_m}}\\ 
    &+\sum_{m=M_s+1}^{2M_s}
    \lVert \hat{{\bm{z}}}_m - h(\bm{x}_b,\bm{l}_m)\rVert^2_{\Sigma_{b_m}}  \\ 
    &+ \lVert \hat{{\bm{x}}}_b - {\bm{x}}_b\rVert^2_{\Sigma_b} + \phi_x(\bm{x}_a) + \sum_{m=1}^{M_s}\phi_l(\bm{l}_m).
\end{split}\label{eq:two-view-map-with-prior}
\end{align}
Here, we use the notation $\lVert x \rVert^2_{\Sigma}\coloneqq x^T \Sigma^{-1}x$ to denote
Mahalanobis distance. $h(\cdot)$ is the measurement model in Eq.~\ref{eq:KMM}, $\hat{{\bm{x}}}_b$ is calculated from the DR data and $\phi_x(\bm{x}_a)$ is the prior on $\bm{x}_a$ whose uncertainty approaches zero, meaning we treat $\bm{x}_a$ as fixed and only adjust $\bm{x}_b$. Note that $\phi_l(\bm{l}_m)$ is the prior on the landmarks obtained from the a-priori map, which is to address the degeneracy illustrated in Fig.~\ref{fig:elevation_degeneracy}.
As we discussed before, both $\bm{x}_b$ and $\bm{l}_m$ are unknown and they can be simultaneously adjusted to satisfy the constraints if without any priors on the landmarks, which means there are no unique solutions. To tackle this degeneracy, we propose to incorporate the information from sidescan data by employing the neural-rendering based techniques outlined in Section~\ref{sec:NR} to construct an initial estimate of the bathymetry as a prior, providing an elevation prior for the landmarks, which is critical to solving the optimization robustly. The landmarks are marginalized out and the relative pose estimation from the MAP estimator is going to be added to the pose graph as loop closure constraints for localization.

\subsection{Incremental Smoothing and Mapping}
We use a pose graph formulation for AUV localization, where the only variables are poses, so that we could exploit the sparsity introduced by this formulation and solve the optimization through iSAM~\cite{kaess2012ijrr}.

The joint probability distribution of the AUV poses $X=[{\bm{x}}_1,{\bm{x}}_2,\ldots,{\bm{x}}_N]$, the LC constraints $Z=[{\bm{z}}_1,{\bm{z}}_2,\ldots,{\bm{z}}_M]$ between two poses, and the DR constraints between successive poses $U=[{\bm{u}}_1,{\bm{u}}_2,\ldots,{\bm{u}}_N]$ are given by:
\begin{equation}
    P(X,U,Z)=P({\bm{x}}_0)\prod_{i=1}^{N}P({\bm{x}}_i|{\bm{x}}_{i-1}, {\bm{u}}_i)\prod_{j=1}^{M}P({\bm{x}}_{b_j}|{\bm{x}}_{a_j},{\bm{z}}_j)
\end{equation}
A MAP estimate of the AUV poses attempts to find the most likely ${\bm{x}}$ by solving the optimization incrementally so that we can avoid the large drift over a long time.

\section{Full System Overview}
Finally, we give an overall view of the whole framework, combining the method for SSS SLAM with the elevation prior introduced in Section~\ref{sec:slam} and the bathymetry reconstruction using neural rendering, from Section~\ref{sec:NR}.

\begin{algorithm}
 \KwData{$D=\{I_{i,n},\hat{\bm{x}}_i^0,h_i\}, D_a=\{(\alpha_m, \beta_m, \gamma_m)\} $}
 \KwResult{$\Phi_\theta, \Phi, R, A_i$ }
 \For{$\textrm{iteration}$ $j=0$ to $J$}{
  $G=\emptyset$\\
 ${\Phi_\theta}^j \gets \{I_{i,n},\hat{\bm{x}}_i^j,h_i\}$ using Algorithm~\ref{alg:NR}\\
 $\mathcal{L}^j\triangleq \{\bm{l}^j_{\gamma_m}\} \gets \{\hat{\bm{x}}_i^j\}, {\Phi_\theta}^j, D_a$\\
 \For{loop all pings $i=1$ to $N$}{
 \If{$i<N_b$}{$G \gets$ AddDRedge$(\hat{\bm{x}}_{i-1}^j,\hat{\bm{x}}_i^j)$\\
 Continue}
 $\mathcal{B}=\{i-N_b,\ldots,i\}  \cap \{\beta_m\}$\\
 \If{$\lvert \mathcal{B} \rvert<$ thres1}{
$G \gets$ AddDRedge$(\hat{\bm{x}}_{i-1}^j,\hat{\bm{x}}_i^j)$}
 \Else{
 $\mathcal{A}=\{\alpha_m | \beta_m \in \mathcal{B}\}$\\
  $\mathcal{L}_{\gamma}^j =\{ \bm{l}_{\gamma_m}^j \in \mathcal{L}^j | \beta_m \in \mathcal{B} \}$\\
 $a,b \gets$ center $\mathcal{A},\mathcal{B}$\\
 \For{RANSAC loop $r=0$ to $R$}{
 $\mathcal{L}_{s}^j =\{ \bm{l}_{\gamma_m}^j | \gamma_m \in_R \mathcal{L}_{\gamma}^j \}$ where $\lvert \mathcal{L}_s^j \rvert=M_s$\\
 ${\mathcal{L}_s^j}^\complement = \{\bm{l}_{\gamma_m}^j \in \mathcal{L}_\gamma^j: \bm{l}_{\gamma_m}^j\notin \mathcal{L}_s^j\} $\\
 $e_{\textrm{before}}^r \gets \textrm{tri}_\textrm{err}({\mathcal{L}_s^j}^\complement, \hat{\bm{x}}_a^j, \hat{\bm{x}}_b^j)$\\
 Solve $\bm{x}^{*,r}_{a},\bm{x}^{*,r}_{b}$ using Eq.~\ref{eq:two-view-map-with-prior} \\
 $e_{\textrm{after}}^r \gets \textrm{tri}_\textrm{err}({\mathcal{L}_s^j}^\complement, \bm{x}^{*,r}_a, \bm{x}^{*,r}_b)$\\
 } 
 $r^* = \textrm{argmin} \{e_{\textrm{after}}^r / e_{\textrm{before}}^r\} $\\
 \If{$e_{\textrm{after}}^{r^*} / e_{\textrm{before}}^{r^*} <$ thres2}{  $G \gets$ AddLCedge$(\bm{x}_{a}^{*,r^*},\bm{x}_{b}^{*,r^*})$ }
 } 
update $G$
} 
$\{{\hat{\bm{x}}_i}^{j+1}\} \gets \{{\hat{\bm{x}}_i}^j, D_a, {\Phi_\theta}^j\}$\\
 }
 \Return $\Phi_{\theta}^{J},\Phi^{J},R^{J}, A^{J}, \{\hat{\bm{x}}_i^J\}$
 \caption{NeuRSS framework}\label{alg:full-system}
\end{algorithm}
We assume data association $D_a = \{(\alpha_m, \beta_m, \gamma_m)\}_{m=1}^{M}$, is known, where the measurement of landmark $\bm{l}_{\gamma_m}$ is obtained from sensor state $\bm{x}_{\alpha_m}$ and $\bm{x}_{\beta_m}$.
Algorithm~\ref{alg:full-system} outlines the NeuRSS framework given SSS, altimeter readings, DR $\{\hat{\bm{x}}_i^0\}$ and data association $D_a$, which can be run iteratively to improve navigation estimates, represented by a pose graph $G$ and the bathymetry estimate, represented by a SIREN $\Phi_\theta$. Line $3$ trains a SIREN given the current navigation estimate using Algorithm~\ref{alg:NR}, where the resultant estimated bathymetry $\Phi_\theta^j$ is used to compute the landmarks 3D positions $\mathcal{L}^j$ given $D_a$ in line $4$. Line $5-25$ describes the iSAM algorithm to estimate the vehicle's poses at all pings' timestamps ($i=1$ to $N$) using $\Phi_\theta^j$ as the elevation prior. Specifically, the pose graph $G$ is constructed at every ping using DR constraints between the two consecutive pings (line $7$ and $11$). At the same time, if the submap $S_b$ with size $N_b$ ending at the current ping $i$ has more than $thres1$ landmarks $\mathcal{L}_\gamma^j$, which can also be observed from a previous submap $S_a$ with size $N_a$, a loop closure is triggered (line $12-24$). Lines $16-22$ describe that the relative pose from the center of $S_a$ to the center of $S_b$ is estimated using RANSAC, where for each iteration $r$, $M_s$ landmarks are randomly sampled from $\mathcal{L}_\gamma^j$ (line $17$), i.e., $\mathcal{L}_s^j$. Given the initial estimate of $\hat{\bm{x}}_a^j,\hat{\bm{x}}_b^j$ before solving the NLS optimization, we can compute the triangulation error $e^r_{\textrm{before}}$ on ${\mathcal{L}_s^j}^\complement$ (lines $18-19$). Line 20 solves the NLS optimization using Eq.~\ref{eq:two-view-map-with-prior} with the elevation prior $\phi_l$ using 
the Levenberg-Marquardt (LM) algorithm to get the optimal estimate $\bm{x}_a^{*,r},\bm{x}_b^{*,r}$, which are then used to compute the triangulation error $e^r_{\textrm{after}}$. After a fixed iterations of RANSAC, if the estimated pose $\bm{x}_a^{*,r^*},\bm{x}_b^{*,r^*}$ that gives the smallest $e^r_{\textrm{after}}/e^r_{\textrm{before}}$, is smaller than a threshold $0<thres2<1$, this relative pose estimate is added to $G$, line $24$. $G$ is updated for every ping (line $25$) until the end of the survey to obtain the pose estimates for the full survey, $\{\hat{\bm{x}}_i^{j+1}\}$, line $26$. Lines $2-26$ can be run several iterations until both of the navigation estimate $\{\hat{\bm{x}}_i^{J}\}$ and bathymetry estimate $\Phi_\theta^J$ are converged.

\section{Experiments}
\subsection{Datasets and Vehicles}
\begin{table}[]\caption{DATASETS DETAILS}\label{tab:datasets}
\begin{tabular}{@{}llll@{}}
\toprule
Survey         & 1           & 2           \\ \midrule
Vehicle            & MMT Ping        & Hugin    \\
Avg altitude (m) & $\sim$17 & $\sim$19 \\
Duration       & 3 hr           & 12 min         \\
Trajectory (km)  & 7.3 & 0.7  \\ 
Survey area       & $\sim$350m$\times$300m &  $\sim$300m$\times$250m      \\
Bathymetry depth range (m)       & 9-25   &  78-92           \\
MBES Bathymetry resolution (m)        & 0.5  &     1         \\
Number of sidescan pings       &  58217            & 3350               \\
Sidescan range (m)     & $\sim$50         & $\sim$170     \\
Sidescan frequency (kHz) & 850     & 410       \\ \bottomrule
\end{tabular}
\end{table}

\begin{figure}[!t]
\includegraphics[width=\linewidth]{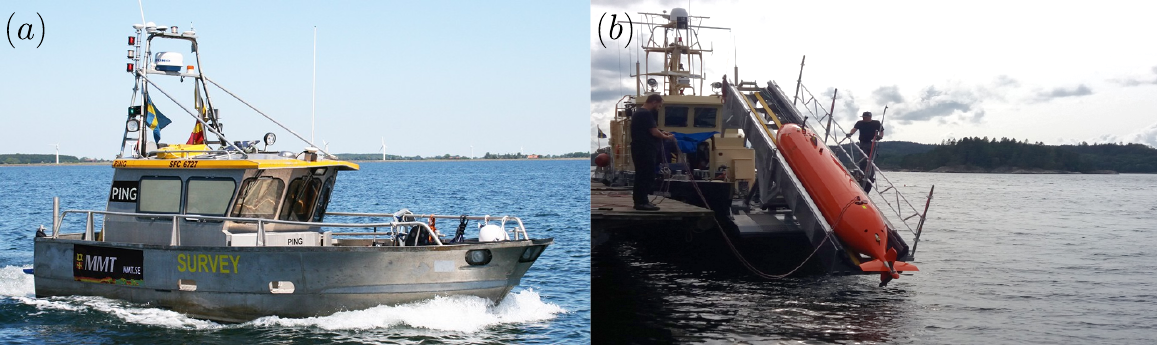} 
  \caption{MMT survey vessel Ping (a) and Hugin AUV (b).}
  \label{fig:AUVs}
\end{figure}
Two datasets have been collected with two vehicles (see Fig.~\ref{fig:AUVs}) for testing the proposed approach. A nearshore surface vessel \textit{MMT Ping} equipped with a real-time kinematic (RTK) GPS, a hull-mounted EdgeTech 4200 sidescan and a Reson SeaBat 7125 MBES has collected Dataset 1, where the surveyed area contains a large mountain and a ridge. Dataset 2 was collected by a Kongsberg Hugin 3000 AUV equipped with a Honeywell HG9900 inertial navigation system (INS), aided with a Nortek 500 kHz Doppler velocity log (DVL), an EdgeTech 2205 sidescan and a Kongsberg EM2040 MBES, where the surveyed terrain is relatively benign with a sinkhole on the seabed. 
Dataset 1 was collected on the surface so that we could have GPS for accurate positioning. The MBES data are used as the ground truth bathymetry during evaluation. Dataset 2 was collected as a part of a long mission (24 hours) without any external navigation aid, where the DR of Hugin will have inevitable drift. However, since Dataset 2 is only 12 minutes and given the high quality of Hugin's INS system we reasonably assume the relative trajectory error is very little within Dataset 2.  Therefore, the DR data and MBES collected are used as the ground truth for evaluation. The main characteristics of the two datasets are summarized in Table \ref{tab:datasets}.

\subsection{SIREN Training Details}
In this work, we leave the architecture of SIREN as it is in~\cite{xie_neural_normal}, a 
5-layer MLP with hidden layer size 128. As for the sidescan, we downsample all the raw data to 512 bins from 5734 bins (Dataset 1) and 20816 bins (Dataset 2) where all 512 bins are used in the loss calculation. All positional data is normalized to [-1,1]. For SIREN's initialization, instead of initializing the weights randomly as in~\cite{xie_neural_normal}, we first train SIREN for 10 epochs using the heightmap linearly interpolated from the sparse altimeter readings, so that the training later could converge faster. 
As for the training hyperparameters, we train 800 epochs using Adam optimizer with a learning rate $2\times10^{-4}$ that linearly decays by a factor of $0.995$ every 2 epochs. For each mini-batch, we randomly selected 200 SSS pings and 800 altimeter readings. For the GD, we optimize 30 steps with $\lambda=2.0$. To compute the transmittance $T$, we sample 30 points along the ray backwards 2 m.

\subsection{Evaluation of the NeuRSS framework}
In this section, we seek to assess and validate the amenability of the proposed NeuRSS framework on large industrial-scale surveys. For this, we have designed \textcolor{black}{two sets of experiments}.  For both experiment setups, we corrupt the ground truth trajectory by adding Gaussian noise to the yaw of the vehicle, $5e-3 rad/s$, to simulate the vehicle's navigation estimates (DR) that inherent cumulative drift. As for the front-end, we generate the ``perfect data association'' using ground truth trajectory and bathymetry. We first construct SSS submaps every 200 pings along the trajectory, and for two submaps that have sufficient overlaps (usually from sections of two parallel adjacent track lines), we use SIFT features to extract feature points from one SSS submap, the reference frame. Then we associate the feature points in the reference waterfall image to their 3D landmarks coordinates using ground truth bathymetry (sidescan draping~\cite{Bore2020}) and then projected them back to the other SSS submap, the current frame. 
\begin{table}[ht!]
\caption{SLAM ATE (m)}
\label{tab:slam}
\centering
\resizebox{\linewidth}{!}{
\begin{tabular}{|ccccc|}
\hline
          & $\{\hat{\bm{x}}_i^0\}$ DR    & $\{\hat{\bm{x}}_i^1\}$ No Prior & $\{\hat{\bm{x}}_i^1\}$ Linear & $\{\hat{\bm{x}}_i^1\}$ SIREN  \\ \hline
Dataset 1 & 9.751 &    8.563     & 6.232                & \textbf{2.074} \\ \hline
Dataset 2 & 7.346 & 5.349   & 2.551                & \textbf{2.060} \\ \hline
\end{tabular}}
\end{table}

\begin{table*}[t]
\caption{RESULTS with SIREN}
\label{tab:slam-iterative}
\resizebox{\linewidth}{!}{
\begin{tabular}{|cccc|cccccc|}
\hline
          & $\{\hat{\bm{x}}_i^0\}$ (DR) ATE & $\{\hat{\bm{x}}_i^1\}$ ATE  & $\{\hat{\bm{x}}_i^2\}$ ATE & $\Phi_\theta^0$ Err. & $\Phi_\theta^0$ Abs. Err. & $\Phi_\theta^1$ Err. & $\Phi_\theta^1$ Abs. Err. & $\Phi_\theta^2$ Err. & $\Phi_\theta^2$ Abs. Err.  \\ \hline
Dataset 1 & 9.751   & 2.074     & 0.822   & -0.037$\pm$0.259 & 0.124 & -0.044$\pm$0.161 &  0.082 & -0.025$\pm$0.155 &  0.069 \\ \hline
Dataset 2 & 7.346   &  2.060    & 1.669   & 0.152$\pm$0.536 & 0.373  & 0.139$\pm$0.498 & 0.287 & 0.137$\pm$0.492 & 0.284\\ \hline
\end{tabular}
}
\end{table*}
In Experiment 1, we have gauged the effects of the landmark elevation prior in the back-end NLS optimization and the SLAM performance on both datasets. We run the NLS optimization on the corrupted trajectory with three setups: no landmark elevation prior, an elevation prior provided using linear interpolation between the altimeter readings of reference frame and current frame, and the elevation prior given by the estimated SIREN bathymetry from SSS using our neural rendering approach. The results of the optimization have been compared based on two error metrics, relative translation error (RTE) translation and absolute trajectory error (ATE), against the ground truth.
\begin{figure}[t]
\includegraphics[width=\linewidth]{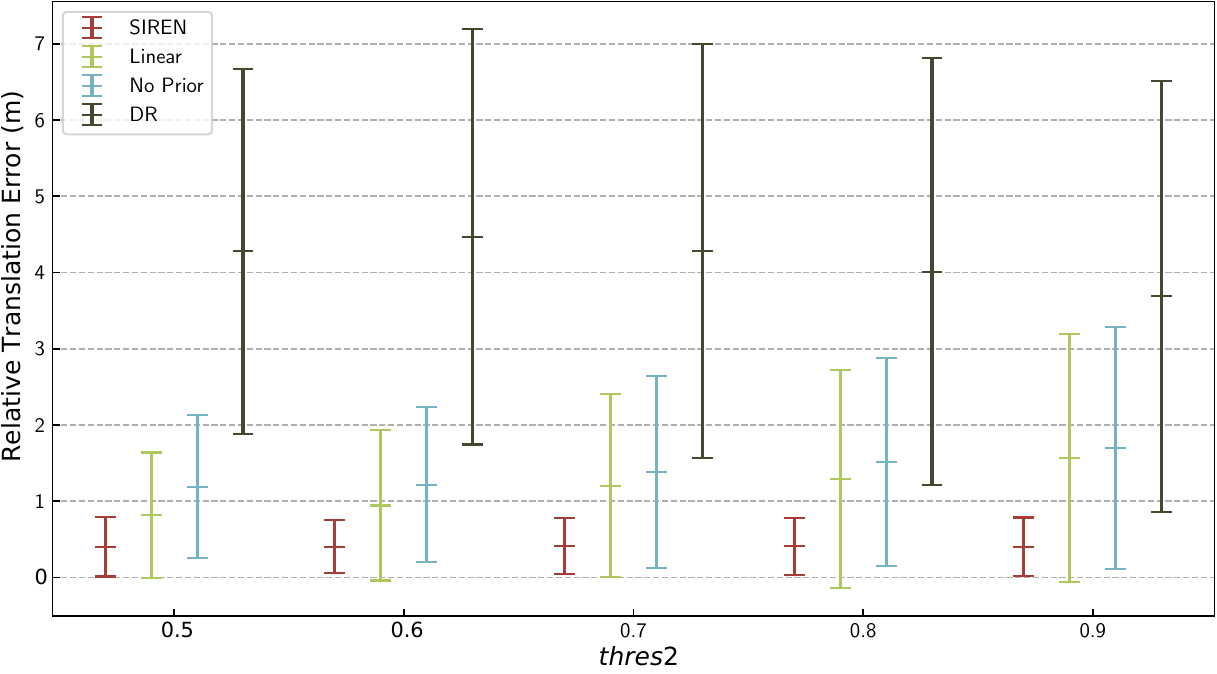}  
  \caption{Mean and standard deviation of the RTE at $thres2=0.5, 0.6, 0.7, 0.8, 0.9$. Black bars are the DR errors before optimization. Red bars and green bars are errors after optimization using SIREN estimates and linear interpolation of altimeter readings as the elevation prior, respectively. Blue bars are the errors without using any elevation priors.}
  \label{fig:elevation_prior}
\end{figure}

In Experiment 2, we have demonstrated how the proposed approach can be used iteratively to improve navigation estimates and subsequently for bathymetric mapping with SSS. On both datasets, we  have applied $J=2$ iterations of Algorithm~\ref{alg:full-system} starting from the corrupted trajectory ($\{\hat{\bm{x}}_i^0\}$, DR). We compare the estimated trajectories and bathymetric maps against their ground truth, respectively.

\section{Results}
\subsection{Exp 1: Elevation Prior on NLS Optimization}

As mentioned in Algorithm~\ref{alg:full-system}, line $24$, a threshold $thres2$ is used to determine if the relative pose estimation from NLS optimization should be added to the pose graph $G$ as LC constraints. This parameter controls whether a LC constraint is considered to have a robust relative pose estimate after optimization. 
Fig.~\ref{fig:elevation_prior} shows the RTE with different $thres2$ for all submap pairs from Dataset 1 after NLS optimization using no elevation prior, linear interpolated elevation prior or the elevation prior from the estimated SIREN bathymetry from SSS. We can observe that the relative pose estimation is not robust at all when no elevation prior is provided, indicating the elevation degeneracy case. Note that we obtain much more robust performance on relative pose estimation when we use Algorithm~\ref{alg:NR} to estimate bathymetry and treat it as a map known a-priori, providing the elevation prior. This indicates that the converged bathymetry estimated from SSS using neural rendering, though having errors due to the inaccurate positioning, still provides a more valuable approximation compared to simply interpolating all the altimeter readings, due to the rich information extracted from SSS imagery.

Table~\ref{tab:slam} shows the ATE when we run the full system (for one iteration) using both datasets, where we can also observe that using the proposed method as the elevation prior gives the smallest trajectory error in both cases. Note that the terrain in Dataset 1 is much more complex than Dataset 2, which is the main reason that the performance using linear interpolation to provide the elevation prior is much worse than that using our estimated SIREN bathymetry on Dataset 1 (6.232 m versus 2.074 m), compared to Dataset 2 (2.551 m versus 2.060 m).

\subsection{Exp 2: Iterative Refinement of Navigation and Mapping}
\begin{figure}[h!]
\centering
\includegraphics[width=\linewidth]{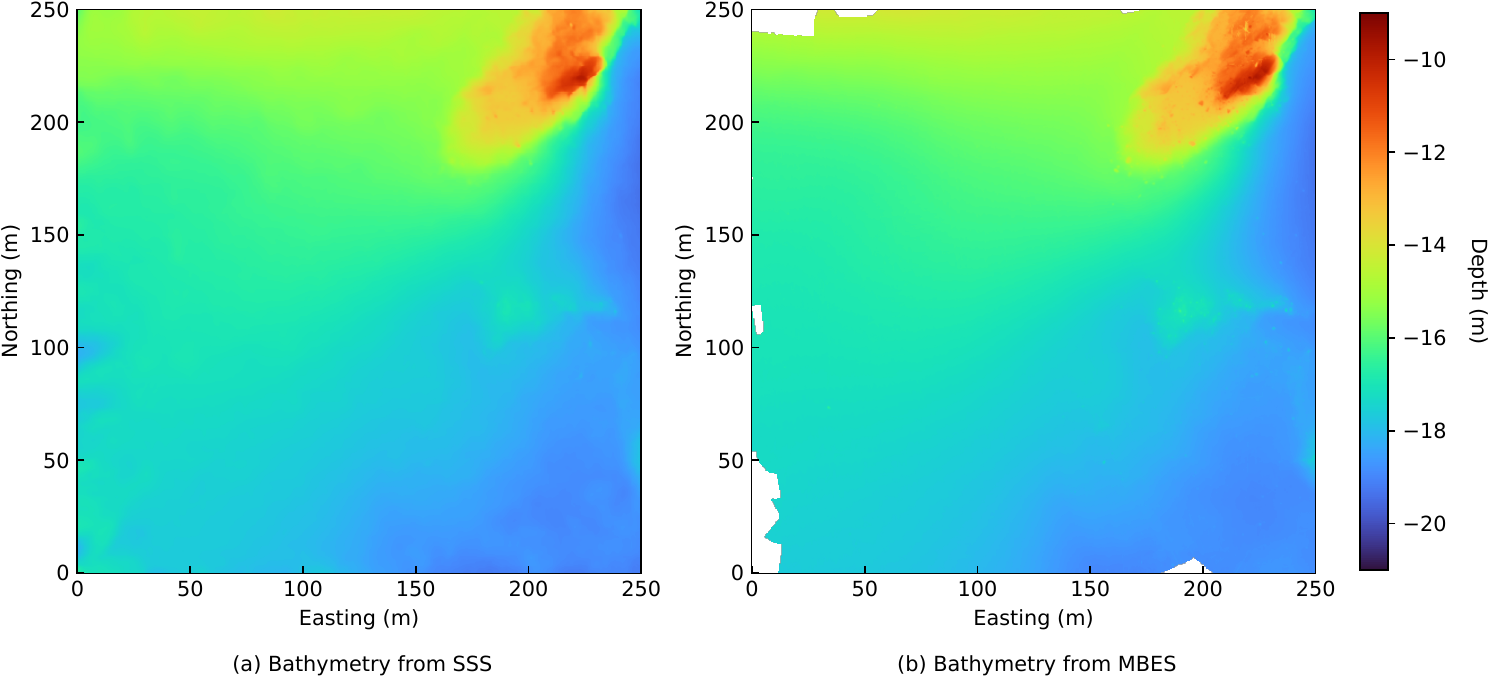} %
  \caption{Final estimated bathymetry from Dataset 1.}
  \label{fig:bathy-1}
\end{figure}

\begin{figure}[h]
\centering
\includegraphics[width=\linewidth]{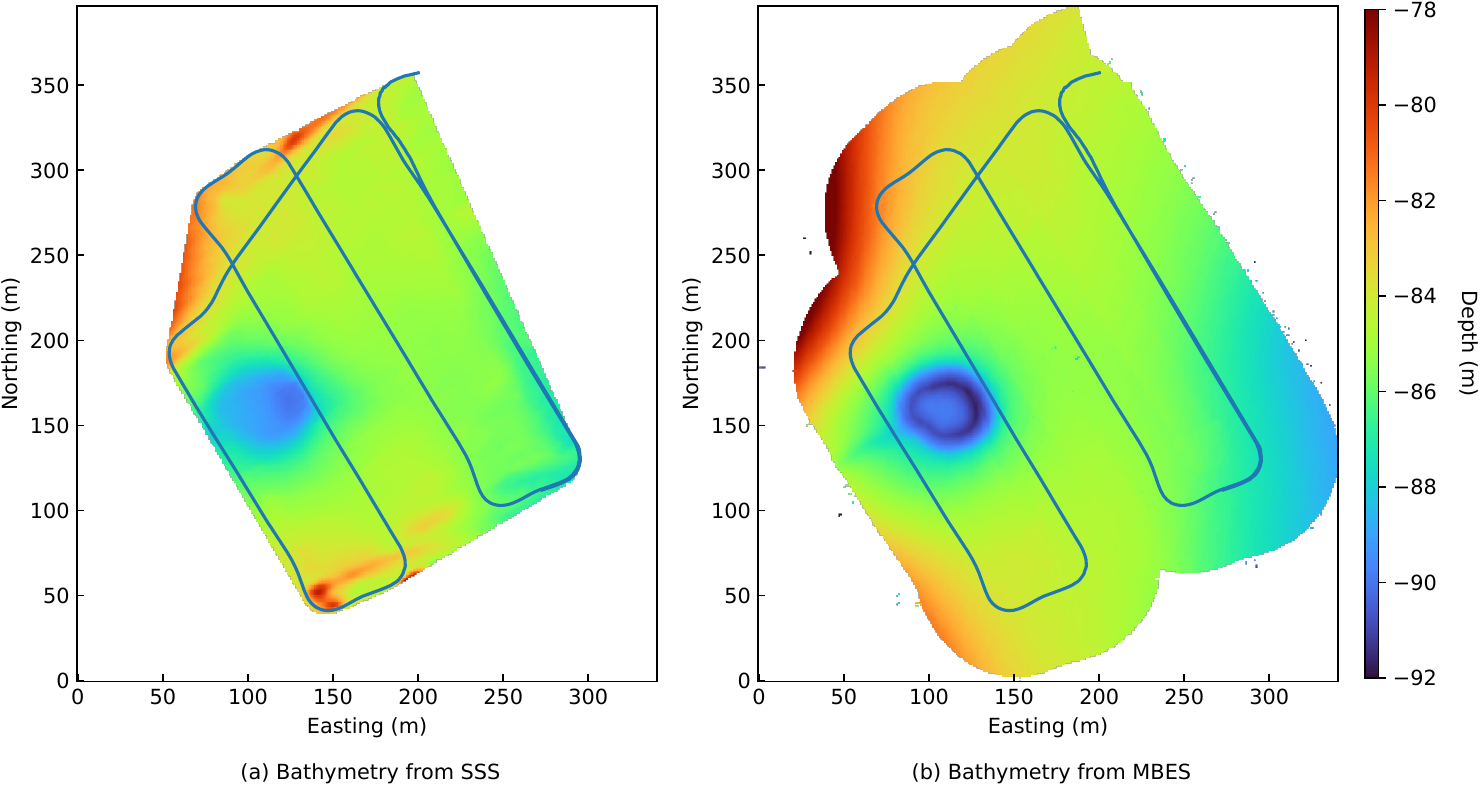} 
  \caption{Final estimated bathymetry from Dataset 2 (AUV trajectory in blue). Note that we use the SSS trajectory to create a boarder of the reconstructed area because the quality outside of the boarder degrades fast due to under-constraints. }
  \label{fig:bathy-2}
\end{figure}

\begin{figure*}[h]
\includegraphics[width=\linewidth]{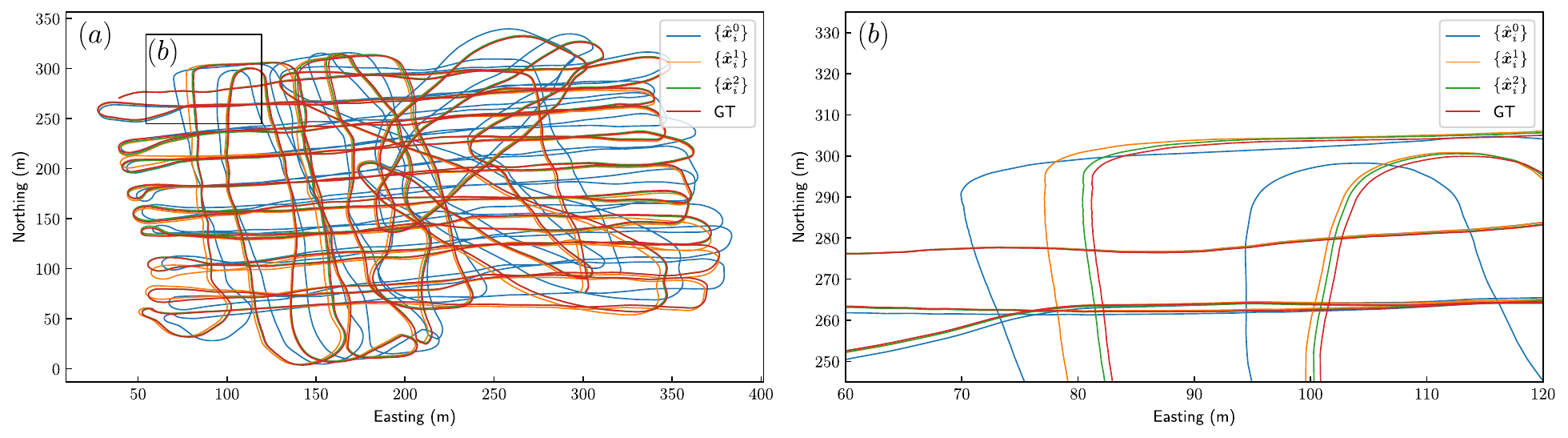} 
  \caption{The optimized trajectory estimates of the vehicle (orange and green), DR estimates (blue), as well as the ground truth for Mission 1. }
  \label{fig:ate1}
\end{figure*}
Table~\ref{tab:slam-iterative}  shows the SLAM ATE  (m) and the error (m) on reconstructed bathymetry $\Phi_\theta^j$ at each iteration, $j=0,1,2$. Inspecting the ATE, on 
 both datasets we can observe that  another iteration ($j=2$)
could further improve the trajectory estimates but again the improvement on Dataset 2 is less than that on Dataset 1, due to the terrain in Dataset 2 is relatively benign and the features on the seabed mainly are clustered around the sinkhole. We can also notice that the errors on the estimated bathymetry in Table~\ref{tab:slam-iterative} also decrease each iteration because of the better estimates on the positioning of the SSS, resulting a final 0.069 m MAE on Dataset 1 and 0.284 m MAE on Dataset 2. Fig.~\ref{fig:ate1} shows the estimated trajectory compared against the ground truth (red) for Dataset 1, the entire mission in (a) and zoom-in section in (b), where it illustrates how our proposed approach can iteratively improve the navigation estimates from DR ($\{\hat{\bm{x}}_i^0\}$, blue) with ATE 9.751 m to $\{\hat{\bm{x}}_i^1\}$ (orange) with ATE 2.074 m and $\{\hat{\bm{x}}_i^2\}$ (green) with ATE 0.822 m.

\begin{figure*}[h]
\centering
\includegraphics[width=\linewidth]{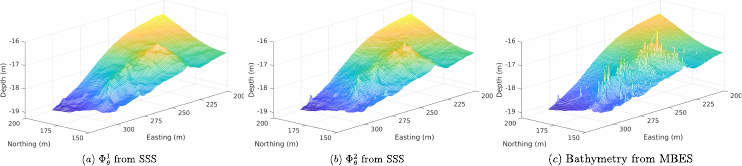}  
  \caption{Zoomed in sections of bathymetry in Dataset 1.}
  \label{fig:bathy_zoomin}
\end{figure*}

Fig.~\ref{fig:bathy-1} shows heightmaps of the final estimated bathymetry from SSS, $\Phi_\theta^2$, and the ground truth bathymetry constructed from MBES, for Dataset 1. We can see that the topographic details of the mountain and the ridge are fairly well reconstructed. We zoom in and show the ridge in 3D as a mesh in Fig.~\ref{fig:bathy_zoomin}, where one can notice the effect of sonar pose estimates on the quality of the reconstructed topology. We can observe that, $\Phi_\theta^2$, in Fig.~\ref{fig:bathy_zoomin} (b) manages to reconstruct more details on the abyssal hills on top of the ridge compared to $\Phi_\theta^1$ in (a). However, compared to MBES in (c), we can see  the limits of the propose approach,
even though the MAE over the whole surveyed area is less than 10 cm.

For Dataset 2, the final estimated bathymetry is shown in Fig.~\ref{fig:bathy-2} (a) with AUV transit in blue, together with the ground truth in (b). Note that the sinkhole can be clearly seen in the estimated bathymetry but the shape and dimension have noticeable errors, compared to the bathymetric map from MBES. This is largely due to the fact that reconstructing bathymetry from SSS is an ill-posed optimization problem, especially in this case, where we lack sufficient repeated observations from different viewpoints (unlike the case in Dataset 1, see Fig.~\ref{fig:ate1}). 
For similar reason, we can also see that the quality of the reconstructed bathymetric map deters at the perimeter.
However, one can reasonably speculate that the reconstruction quality could improve given more survey lines, for example perpendicular to the ones in Fig.~\ref{fig:bathy-2}.

\section{CONCLUSIONS and future work}
We have presented NeuRSS, a neural rendering-based framework for reconstructing bathymetry from SSS data and DR estimates in a self-supervising manner. The proposed framework has been tested on two field datasets collected with different robots. We demonstrated that the bathymetric estimates from SSS using neural rendering play an important role in addressing the elevation degeneracy in the NLS optimization for estimating the relative poses between two submaps. Compared to interpolating between altimeter readings, the elevation prior provided by incorporating SSS with neural rendering, results in much more robust optimization, especially when the terrain is complex, as in Dataset 1.
We also show that the proposed approach can be run iteratively to improve navigation and bathymetric estimates for high-quality bathymetric mapping using SSS data from standard surveys.

The current major limitation of this work is that we did not address the data association problem in the front-end of SSS SLAM pipeline. Automatic data association in SSS imagery is still an active and open research question, largely due to the unique challenges that come from the special sensor modality of SSS. In~\cite{zhang2023ietrsn,zhang2023dense}, canonical transformation under flat seafloor assumption has been applied to SSS images to reduce geometric and radiometric distortions before feature-based matching~\cite{zhang2023ietrsn} and dense matching~\cite{zhang2023dense}. One possible future work is to incorporate the bathymetric estimates from neural rendering into SSS canonical transformation and apply similar matching approaches for automatic data association.

Another possible future work is to use MBES and SSS data for AUV SLAM and super-resolution bathymetric mapping in a neural rendering framework, leveraging the strengths of both sensors, namely MBES's ability to direct measure the 3D seafloor geometry and SSS's wide swath range and high-resolution imagery.

Another limitation is that our method is mostly suited to offline optimization instead of real-time applications. The main reason is that to train a SIREN to converge to a self-consistent bathymetric map with high quality and fidelity, we need the multiple repeated observations from SSS from different viewpoints. Nevertheless, if one can explore the idea of combining MBES and SSS at the same time, it is possible to run dense SLAM in an embedded platform in real time.






\section*{ACKNOWLEDGMENT}
The authors thank the Alice Wallenberg Foundation for funding MUST, Mobile Underwater System Tools, Project that provided the Hugin AUV. The computations were enabled by resources provided by the National Academic Infrastructure for Supercomputing in Sweden (NAISS) and the Swedish National Infrastructure for Computing (SNIC) at Berzelius partially funded by the Swedish Research Council through grant agreements no. 2022-06725 and no. 2018-05973.


\bibliographystyle{IEEEtran}
\bibliography{IEEEabrv,root}

\end{document}